\documentclass[conference]{IEEEtran}  
\IEEEoverridecommandlockouts
\usepackage{tabularx}
\usepackage{amsmath,amsfonts}

\usepackage{threeparttable}
\usepackage{cite}
\usepackage{multirow}
\usepackage{longtable}
\usepackage{graphicx}
\usepackage{caption}
\usepackage{subcaption}
\usepackage{float}

\usepackage{soul}
\usepackage{url}
\usepackage{booktabs}
\usepackage{algorithm}
\usepackage{algorithmic}
\usepackage[switch]{lineno}

\usepackage{lineno}
\usepackage[hidelinks]{hyperref}

\hypersetup
{
 unicode=false,     
 pdftoolbar=true,    
 pdfmenubar=true,    
 pdffitwindow=false,   
 pdfstartview={FitH},  
 pdftitle={My title},  
 pdfauthor={Author},   
 pdfsubject={Subject},  
 pdfcreator={Creator},  
 pdfproducer={Producer}, 
 pdfkeywords={keywords}, 
 pdfnewwindow=true,   
 colorlinks=true,    
 linkcolor=red,     
 citecolor=blue,    
 filecolor=magenta,   
 urlcolor=cyan      
}

\def\BibTeX{{\rm B\kern-.05em{\sc i\kern-.025em b}\kern-.08em
    T\kern-.1667em\lower.7ex\hbox{E}\kern-.125emX}}

\title{HONEST-CAV: Hierarchical Optimization of Network Signals and Trajectories for Connected and Automated Vehicles with Multi-Agent Reinforcement Learning}

\author{Ziyan Zhang$^{*,\dagger}$, Changxin Wan$^{\dagger}$, Peng Hao,~\IEEEmembership{Member,~IEEE}, Kanok Boriboonsomsin,~\IEEEmembership{Member,~IEEE},\\ Matthew~J.~Barth,~\IEEEmembership{Fellow,~IEEE}, Yongkang Liu, Seyhan Ucar, Guoyuan~Wu,~\IEEEmembership{Senior~Member,~IEEE}
\thanks{The work is funded by Toyota InfoTech Labs.}
\thanks{Ziyan Zhang, Changxin Wan, Peng Hao, Guoyuan Wu, Kanok Boriboonsomsin, and Matthew J. Barth are with the College of Engineering, Center for Environmental Research and Technology, University of California at Riverside, Riverside, CA, 92507.}
\thanks{Yongkang Liu and Seyhan Ucar are with the InfoTech Labs, Toyota Motor North America, Mountain View, CA, 94043.}
\thanks{$^{*}$Corresponding author. e-mail: ziyan.zhang@email.ucr.edu}
\thanks{$^{\dagger}$Ziyan Zhang and Changxin Wan contributed equally to this work.}
\thanks{\textbf{Accepted at IV 2026. Final version to appear in IEEE Xplore.}}
}

\begin{document}

\maketitle
\begin{abstract}
This study presents a hierarchical, network-level traffic flow control framework for mixed traffic consisting of Human-driven Vehicles (HVs), Connected and Automated Vehicles (CAVs). The framework jointly optimizes vehicle-level eco-driving behaviors and intersection-level traffic signal control to enhance overall network efficiency and decrease energy consumption. A decentralized Multi-Agent Reinforcement Learning (MARL) approach by Value Decomposition Network (VDN) manages cycle-based traffic signal control (TSC) at intersections, while an innovative Signal Phase and Timing (SPaT) prediction method integrates a Machine Learning-based Trajectory Planning Algorithm (MLTPA) to guide CAVs in executing Eco-Approach and Departure (EAD) maneuvers. The framework is evaluated across varying CAV proportions and powertrain types to assess its effects on mobility and energy performance.
Experimental results conducted in a 4×4 real-world network demonstrate that the MARL-based TSC method outperforms the baseline model (i.e., Webster method) in speed, fuel consumption, and idling time. In addition, with MLTPA, HONEST-CAV benefits the traffic system further in energy consumption and idling time. With a 60\% CAV proportion, vehicle average speed, fuel consumption, and idling time can be improved/saved by 7.67\%, 10.23\%, and 45.83\% compared with the baseline. Furthermore, discussions on CAV proportions and powertrain types are conducted to quantify the performance of the proposed method with the impact of automation and electrification.
\end{abstract}


\section{Introduction}

Due to the rapid growth of traffic demand and fuel usage in urban cities, traffic operation and management methods, such as vehicle trajectory planning and intersection adaptive signal control, have become increasingly critical for traffic efficiency improvement and energy savings  \cite{serok2022identification, davis2021transportation, hussain2023investigating}. However, conventional methods still treat trajectory planning and traffic signal control as separate problems, limiting the potential benefits of connected mobility, especially under mixed traffic conditions \cite{zhou2024cooperative, almannaa2019field, esaid2021machine}. Connected and Automated Vehicles (CAVs), enabled by Vehicle-to-Infrastructure (V2I) communication technology, offer new opportunities for the collaboration of vehicle trajectory planning and traffic signal control. 

The vehicle-signal cooperative control has drawn sustained attention since 2010\cite{malakorn2010assessment}. Early works demonstrated the promise of cooperative signal–vehicle control but were mainly limited to isolated intersections by rule-based methods\cite{malakorn2010assessment, li2014signal}. Later formulations extended to mixed-integer programming \cite{yu2018integrated, tajalli2021traffic} to achieve global optima, while it is difficult to scale up due to the highly computational burden. The learning-based method, such as imitation learning (IL) for vehicles \cite{esaid2021machine} and multi-agent reinforcement learning (MARL) for intersections \cite{mei2023libsignal}, has been integrated by a hierarchical framework to achieve vehicle-signal cooperative control while considering computational efficiency.

Overall, existing co-optimization approaches remain constrained by computational burden and limited scalability. Recent research trends emphasize learning-based methods with a hierarchical framework capable of handling mixed traffic and broader network structures with computational efficiency. Moreover, an Internal Combustion Engine Vehicle (ICEV) powertrain type is basically assumed in existing methods, while the performance with different powertrain types (i.e., electric vehicle) is rarely considered.

In this study, we propose a hierarchical framework based on MARL and IL to address the co-optimization of traffic control in large-scale networks with CAVs. At the intersection level, the MARL with the value decomposition network (VDN) is applied for traffic signal control. At the vehicle control layer, we incorporate an IL-based Eco-Approach and Departure (EAD) strategy \cite{esaid2021machine}. Moreover, a novel green phase prediction algorithm is proposed to facilitate coordination between the two layers. The proposed framework is modular and scalable, allowing for the replacement or integration of different traffic signal control and EAD methods. With trained MARL-based signal control and the vehicle-level EAD strategy, the framework demonstrates strong potential for real-time, scalable deployment in larger urban networks.

The main contributions of this paper are as follows: 

(1) Proposing a hierarchical co-optimization framework with high scalability and real-time efficiency for signal and vehicle control at the network level, enabling coordinated decision-making across multiple intersections and vehicles.

(2) Introducing a MARL-based traffic signal control approach enhanced with multi-processing, which significantly accelerates the training process by simulating multiple traffic scenarios in parallel, improving scalability and robustness.

(3) Developing a Signal Phase and Timing (SPaT) prediction method based on historical data and current traffic conditions to improve the reliability of EAD strategies, thereby enhancing fuel efficiency and reducing emissions.

(4) Analyzing the performance of the proposed framework under varying CAV/CAEV proportions in different powertrain types, demonstrating its effectiveness, adaptability, and generalization capability across diverse traffic environments.


\section{Problem Description}

\begin{figure*}[htbp]
  \centering
  \includegraphics[width=0.95\linewidth]{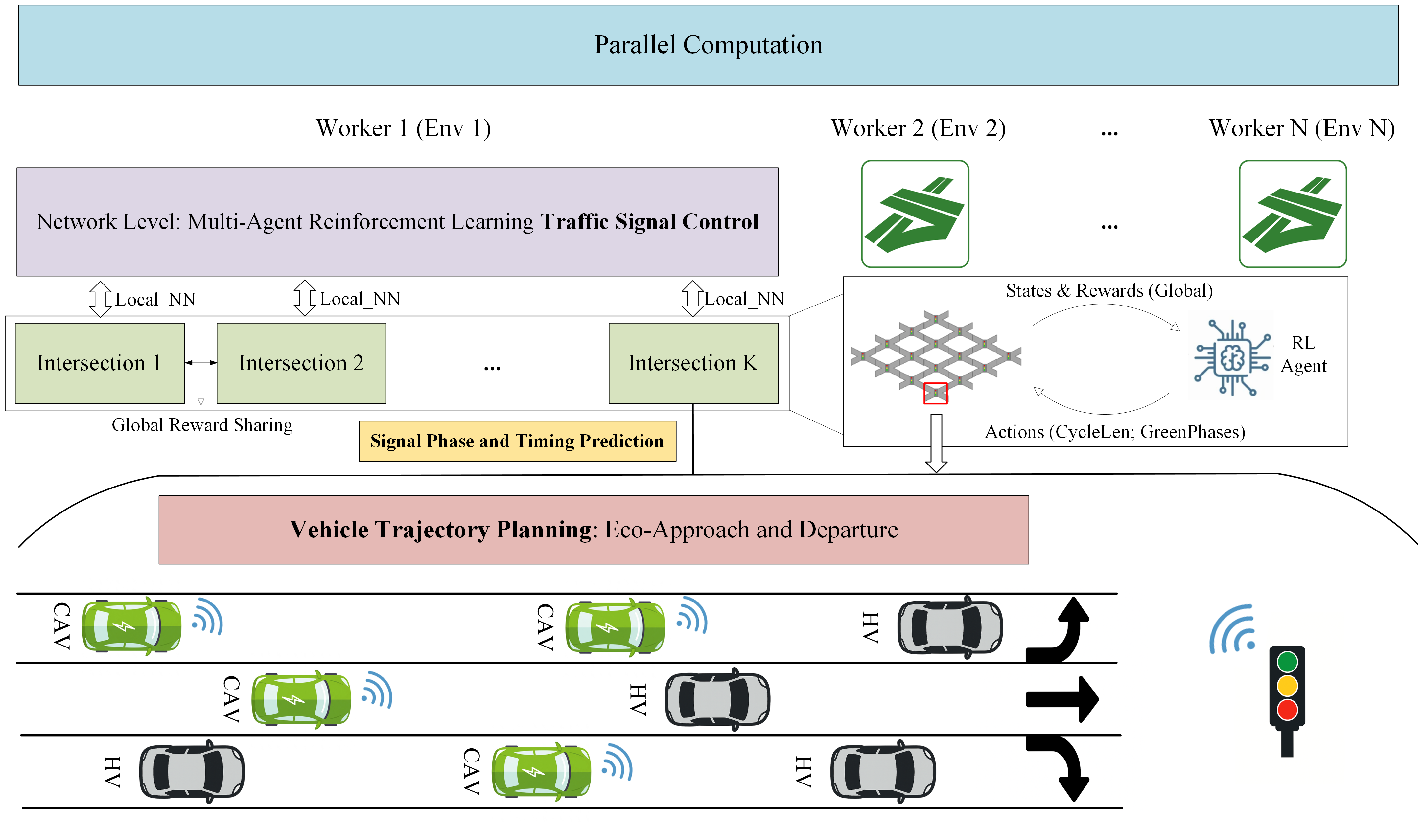}
  \caption{The Schematic Diagram of HONEST-CAV}
  \label{fig:MARL}
\end{figure*}

This study focuses on an urban network consisting of multiple signalized intersections, and two types of vehicles, i.e., HVs and CAVs are considered. The goal of this study is to achieve a global minimum of both the overall network's efficiency and energy consumption by dynamically adjusting the SPaT settings of the intersections and regulating the driving behaviors of CAVs. However, it is challenging to obtain such a global optimum due to the complicated interactions between the signal control and vehicle trajectory planning. Instead, a hierarchical mixed traffic flow control framework that decouples this interaction is applied to achieve the global optimal approximately, as illustrated in Fig. ~\ref{fig:MARL}. The goal is now divided into two parts: (i) intersection-level signal control for network efficiency improvement, and (ii) vehicle-level EAD strategy for energy savings. The detailed modules for this framework are as follows:

For intersections, it is assumed that they have uniform channelization and exhibit the capacity to collect real-time traffic flow information (e.g., density, average speed) by roadside units (RSUs), and then dynamically optimize their SPaT actions (i.e., cycle length and green phase durations) to reduce the overall network delay and queue length. The key challenge lies in coordinating the actions of individual intersections to improve overall network efficiency. To address this, we adopt a MARL-based traffic signal control (TSC) framework that incorporates global reward estimation, enabling agents to account for network-level performance with local decisions.

For each CAV, its longitudinal trajectory can be optimized by regulating its speed profile that allows crossing during the green light window $\left[T_{\mathrm{start}}, T_{\mathrm{end}}\right]$ with minimal energy usage, while such a time window is not static under RL-based signal control. Therefore, a prediction mechanism is proposed to estimate the next-cycle passing interval $[T_{\mathrm{start}}^{\mathrm{pred}},\; T_{\mathrm{end}}^{\mathrm{pred}}]$ for real-time vehicle-level trajectory planning.

\section{Methodology}
\subsection{MARL-based Traffic Signal Control}
MARL-based control methods can be generally categorized into (1) centralized training and centralized execution (CTCE) and (2) centralized training and decentralized execution (CTDE) approaches~\cite{hua2025multi}. CTCE can achieve globally optimal signal plans but suffers from high computational cost and latency, making it unsuitable for large-scale real-time deployment. In contrast, CTDE treats each intersection as an autonomous agent, enabling scalable and efficient network-level coordination. Therefore, we adopt a CTDE framework for intersection-level MARL-based TSC.

As illustrated in Fig.~\ref{fig:MARL}, each intersection collects local transitions (state–action–reward), while a global reward is formed by aggregating cycle-based performance across the network. These experiences are jointly used to train a shared policy, which benefits from diverse traffic dynamics and improves convergence and generalization. To further accelerate training, we employ an asynchronous multi-processing scheme, where multiple simulation processes run in parallel, update the shared global policy together~\cite{mnih2016asynchronous}, and periodically update the local policy. Additionally, we adopt a cycle-based control strategy, where signal timing (cycle length and green splits) is updated once per cycle. This provides more stable signal patterns and is particularly compatible with EAD-based vehicle trajectory planning, which relies on predictable green intervals. The objective of the MARL-based algorithm is to learn a policy $\pi$ that maximizes the expected cumulative global reward $R_{\mathrm{global}}$ with entropy regularization:
\begin{equation}
    J(\pi) = \sum_{t}\mathbb{E}_{\boldsymbol{s}_t,\boldsymbol{a}_t} 
    [r_{\mathrm{global}}(\boldsymbol{s}_t, \boldsymbol{a}_t)
    + \alpha \mathcal{H}(\pi(\cdot|\boldsymbol{s}_t))]
\end{equation}
where \( \boldsymbol{s}_{t}=\cup_i{\boldsymbol{s}_{t}^i} \) and \( \boldsymbol{a}_{t}=\cup_i{\boldsymbol{a}_{t}^i} \) denote the global state, action at $t$-th cycle. \( \alpha \) is the temperature coefficient, and 
$\mathcal{H}(\pi(\cdot|\boldsymbol{s}_t))$
denotes the entropy of the policy for better exploration.

\subsubsection{Definition of States}
The state space is constructed to capture the spatio-temporal traffic information at a four-legged intersection over a full signal cycle. We first divide the entire intersection state into 12 lane-based states (i.e., 4 directions multiple 3 lanes), respectively. For each lane-based state, a weighted temporal aggregation method is introduced to capture second traffic states over a full signal cycle while considering three time-weighted average cases. The formulas of the state space \( \boldsymbol{s}_t^i \) for each intersection are as follows:
\begin{equation}\label{eq:real_time feature}
    \left\{
    \begin{aligned}
    &     \boldsymbol{s}_t^i = \{\cup_{d=1}^{4}\cup_{l=1}^{3}\cup_{p=1}^{3}\hat{s}_{t}^i(d, l, p)\}\\ 
    &s_{t}(d, l, p) = \sum_\tau w_\tau^p \cdot o_\tau(d,p)/{\sum_\tau w_\tau^p} \\
    &w_\tau^p = \lambda_p \cdot w_{\tau+1}^p, ~p=0,1,2\\
    &\lambda_0=1.05,\lambda_1=1.0,\lambda_2=0.95
    \end{aligned}
    \right.
\end{equation}
where \( \hat{s}_{t}(d, l, p) \) denotes the \( l\text{-th} \) lane-based state with \( p\text{-th} \) time-weighted average cases at $d$-th direction. $\tau$ means the \( \tau\text{-th} \) time-step at the current phase of each intersection. \(w_{\tau}^p\) represents the importance of \( \tau\text{-th} \) step-level traffic state with \( p\text{-th} \) time-weighted average cases, \(w_{\tau=0}^p=1, p=1,2,3\). \( \lambda_p \) determines the \( p\text{-th} \)  temporal weighting pattern, respectively. 

\(o_\tau(d,p) \) represents the observation collected by RSUs equipped in each lane at time step \( \tau \). In this study, four types of information are considered to be collected, defined by:
\begin{equation}
    o_\tau(d,p) = \{\text{Occ}, \text{AvgSpd}, \text{QLen}, \text{CAVPR}\}
\end{equation}
where \text{Occ}, \text{AvgSpd}, \text{QLen}, and \text{CAVPR} present the traffic flow density, average speed, queue length, and CAV penetration rate within the detection range.

Given 4 directions, 3 lanes per direction and 3 aggregated time states per lane, and 4 traffic features per state, the state input of each intersection agent is a 144-dimensional vector, which provides a comprehensive representation of intersection-level traffic dynamics for downstream RL tasks.

\subsubsection{Definition of Actions} The SPaT setting consists of left and straight phases for west-east and north-south direction pairs. The intersection agent aims to optimize the cycle length and green duration of the 4 phases, so the action space for each agent can be determined as a 5-dimensional vector \( \boldsymbol{a}_t^i = \{u_t^{i}, v_{t,j}^{i}\}, \text{where}~ j=1,\dots,4 \), defined as follows:

\begin{itemize}
    \item \textbf{Cycle Length Change Ratio}~ $u_t^{i} \in [-0.7, 0.7]$: a scalar states the relative change ratio to the base cycle length.
    \item \textbf{Green-to-Cycle Ratio of Each Phase} $v_{t,j}^{i} \in [0, 1]$: the green-to-cycle ratio of the \( i\text{-th} \) phase.
\end{itemize}
To achieve cooperative signal control, all the signal agents are assumed to take action synchronically, so a common cycle length is required for the intersection agents. Moreover, the action for green duration needs to rescale with a summation of 1, and a minimum green time is needed for the intersection to cover the start-up lost time and ensure driving safety. Based on the above rules, the SPaT setting for the next cycle for given an action $\boldsymbol{a}_t^i$ are computed by:
\begin{equation}\label{eq:action transformation}
    \left\{
    \begin{aligned}
    &t_{\mathrm{cyc}} = 
    \frac{1}{N}
    \sum_{i=1}^{N}\mathrm{clip}\!\left( 
        t_{\mathrm{cyc},b}  (1 + u_{t}^i),\ 30,\ 150 
    \right)\\ 
    & t_{\mathrm{gre}}^i = t_{\min} + t_{\mathrm{adj}}\cdot \mathrm{Softmax}\left( v_{t,j}^i\right)\\
    & t_{\mathrm{adj}} = t_{\mathrm{cyc}} - 4\cdot  t_{\mathrm{switch}} - 4\cdot t_{\min}
    \end{aligned}
    \right.
\end{equation}
where \( t_{\mathrm{cyc}}\) denotes the duration of next common cycle length. \( t_{\mathrm{adj}}\) is the adjustable duration for green phases. \(t_{\mathrm{gre}}^i\) is the \( j\text{-th} \) green duration, including W-E left, W-E straight, N-S left, and N-S straight. A switch phase is between two consecutive green phases. \(t_{cyc,b}\), \(t_{\mathrm{switch}}\), and \(t_{\min}\) represent the basic cycle length, switch phase, and the minimum green time duration. \text{clip}\((\cdot,~\min,~\max)\) limits a value to within a specified range. \text{Softmax($\cdot$)} represents the softmax operator.

\subsubsection{Definition of Reward}
Specifically, the global reward is defined as the summation of local agents' reward. The local reward at each decision step (i.e., each cycle) is defined as the accumulated cycle-average passed vehicle numbers and queue length across all directions; the global reward is defined as the average value of local rewards of all agents:
\begin{equation}
    r_t^{\mathrm{global}} = \frac{1}{t_{cyc}}\sum_{i=1}^{N} \sum_{\tau=1}^{t_{\mathrm{cyc}}}(p_i^{\tau}-\omega q_i^{\tau})
\end{equation}
where $p_i^{\tau}$, $q_i^{\tau}$ denote the number of the passed vehicles and queue length for the $i$-th intersection at the $\tau$-th time step during the $t$-th signal cycle. $\omega$ is the weight penalty for queue length, $\omega=12$. Such a reward encourages agent to pass more vehicles with less queue length.

\subsubsection{Multi-Agent Soft Actor-Critic}
A Multi-Agent Soft Actor-Critic (MASAC) algorithm is developed to enable coordinated signal control. Each intersection agent is equipped with a local policy network (actor) and a local value function (critic), while sharing a global reward to achieve cooperative optimization across the network. To accurately estimate the global reward, the global Q-value function is approximated by the aggregation of distributed local Q-functions using a value decomposition network (VDN) \cite{sunehag2017value}:
\begin{equation}
    Q(\boldsymbol{s}_t,\boldsymbol{a}_t) 
    \approx \sum_{i=1}^{N} Q_{i}(\boldsymbol{s}_{t}^i, \boldsymbol{a}_{t}^i|\theta_i)
\end{equation}
where $\theta_i$ is the critic network for $i$-th intersection.

Following the standard SAC principle~\cite{haarnoja2018soft}, each local actor network \( \pi_{\vartheta_i} \) generates an action based on its own local observation \( \boldsymbol{s}_{t}^i \), 
while the global critic aggregates the outputs from all local Q-functions to estimate the overall value.
Accordingly, the actor loss of agent \( i \) is optimized with respect to the global value:
\begin{equation}
    J_{\pi_i}(\vartheta_i) = 
    \mathbb{E}_{\boldsymbol{s}_{t}^i \sim \mathcal{D}, 
               \boldsymbol{a}_{t}^i \sim \pi_{\vartheta_i}}
    \Big[\alpha \log\pi_{\vartheta_i}(\boldsymbol{a}_{t}^i|\boldsymbol{s}_{t}^i) 
    - Q(\boldsymbol{s}_{t}, \boldsymbol{s}_{t})\Big]
\end{equation}
where and \( \mathcal{D} \) denotes the replay buffer.

The critic network minimizes the temporal-difference (TD) error of the global reward using the L2 loss:
\begin{equation}\label{eq:fit global Q}
    \left\{
    \begin{aligned}
    &J_{Q_i}(\theta_i) = \mathbb{E}_{(\boldsymbol{s}_t,\boldsymbol{a}_t,r_t^{\mathrm{global}},\boldsymbol{s}_{t+1}) \sim \mathcal{D}}
    \Big[y_t - Q(\boldsymbol{s}_t,\boldsymbol{a}_t)\Big]^2\\ 
    & y_t = r_t^{\mathrm{global}}
    + \gamma\, Q
    (\boldsymbol{s}_{t+1}, \boldsymbol{a}_{t+1})
    \end{aligned}
    \right.
\end{equation}
where \( \gamma \) is the discount factor and 
\( \boldsymbol{a}_{t+1} \sim \pi(\boldsymbol{s}_{t+1}) \).
During training, a shared actor ($\theta_1=\dots\theta_N=\theta$) and critic network ($~\vartheta_1=\dots\vartheta_N=\vartheta$) is implemented for each intersection for robust training, the critic parameters are softly updated with target networks to enhance stability.

\subsection{Signal Phase and Timing Prediction Algorithm}
Accurate generation of CAV speed profiles requires knowledge of the corresponding green passing interval. However, under RL-based traffic signal control, the SPaT of the next cycle is not determined until the current cycle ends, thus necessitating prediction. Existing SPaT prediction methods in semi-actuated or adaptive signal systems mainly rely on historical phase patterns and often assume relatively stable cycle lengths, limiting their effectiveness under dynamic conditions \cite{genser2024time}. 

\begin{figure}[htbp]
  \centering
    \includegraphics[width=\linewidth]{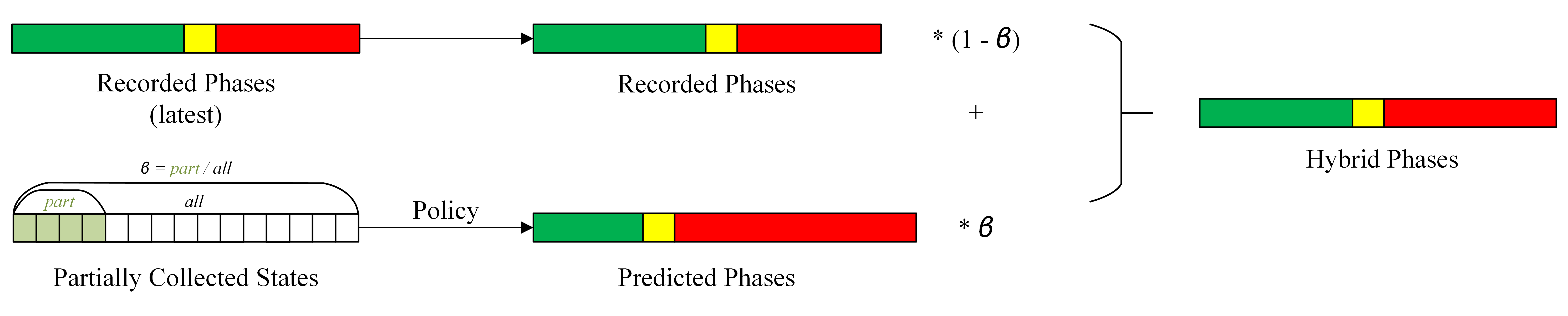}
  \caption{The SPaT Prediction Algorithm}
  \label{fig:greenphasepred}
\end{figure}

In this work, because the signal timing is determined by an adaptive RL policy, we first estimate the upcoming SPaT by executing the trained policy based on currently observed intersection states. However, early-cycle observations may be insufficient for accurate inference. To address this, we propose a hybrid prediction mechanism that blends the policy-driven estimate with historical SPaT patterns using a time-dependent weighting factor, as shown in Fig.~\ref{fig:greenphasepred}. This approach improves robustness when real-time state information is limited.
\begin{equation}\label{eq:fit global Q}
    \left\{
    \begin{aligned}
    &\mathrm{HPhase} = \left\lfloor (1 - \beta) \cdot \mathrm{RPhase} + \beta \cdot \mathrm{PPhase} \right\rfloor\\
    &\beta = \min\left( \frac{\tau}{t_{\mathrm{cyc}}},\ 1.0 \right)
    \end{aligned}
    \right.
\end{equation}
where HPhase, RPhase and PPhase represents hybrid phases, recorded phases and predicted phases, respectively.

\subsection{Imitation Learning Based Vehicle Trajectory Planning}
Regarding vehicle trajectory planning, Eco-Approach and Departure (EAD) strategies generally face a trade-off between computational efficiency and trajectory optimality. Rule-based methods offer real-time execution but often lack precision under dynamic traffic conditions \cite{asadi2009predictive, hao2015developing}. In contrast, optimization-based approaches, such as the Graph-Based Trajectory Planning Algorithm (GBTPA) \cite{hao2021connected}, can produce near-optimal trajectories by minimizing objectives such as energy or acceleration, but their high computational cost limits real-time deployment, especially in large-scale networks \cite{esaid2021machine}.

To address this issue, the Machine Learning Trajectory Planning Algorithm (MLTPA) \cite{esaid2021machine} was developed as a surrogate model trained on GBTPA-generated optimal trajectories. MLTPA imitate the optimal trajectories from GBTPA accuracy while significantly reducing computation, enabling real-time EAD control. Therefore, MLTPA is used in this study to generate the initial speed profile for CAVs.

GBTPA, which forms the basis of MLTPA, constructs a state-space graph where nodes represent vehicle states and edges encode energy-related transition costs. The optimal trajectory is identified via Dijkstra’s algorithm. Algorithm 1 summarizes the corresponding input and output settings.

\begin{algorithm}[H]
\caption{Algorithm for GBTPA}
\begin{algorithmic}[1]
\renewcommand{\algorithmicrequire}{\textbf{Inputs:}}
\renewcommand{\algorithmicensure}{\textbf{Output:}}
\REQUIRE 
PassingIntervalStart\\
\#The earliest time the vehicle can cross the intersection.\\
PassingIntervalEnd\\
\#The latest time the vehicle can cross the intersection.\\
DistanceToSignal\\
\#The distance between the vehicle and the intersection.\\
InstantVelocity\\
\#The instantaneous velocity of the vehicle.\\
\ENSURE  SpeedProfile
\end{algorithmic}
\end{algorithm}

\section{Simulation and Evaluation}

In this section, we first utilized a simulated network to test the performance of our proposed HONEST-CAV, and discussed the impact of CAV proportions and powertrain type to on the proposed hierarchical co-optimization framework.

\subsection{Simulation Settings}

\begin{figure}[htbp]
  \centering
  \includegraphics[width=\linewidth]{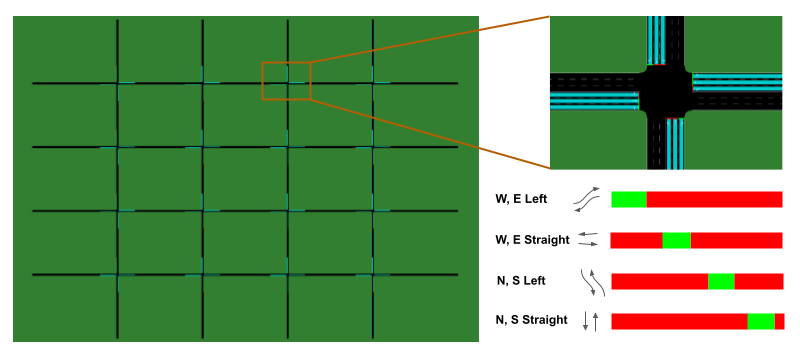}
  \caption{The Environment of Traffic Network}
  \label{fig:roadnetwork}
\end{figure}

The simulation experiments are conducted using the open-source traffic simulator SUMO \cite{lopez2018microscopic}. As shown in Fig.~\ref{fig:roadnetwork}, a real-world $4 \times 4$ urban network in Hangzhou with calibrated traffic demand from CityFlow \cite{wei2019survey} is adopted. The simulation starts with an empty network and terminates after all vehicles have exited. The traffic stream consists of both HVs and CAVs, modeled based on SUMO’s default passenger ICEV configuration. Moreover, multiple random seeds are used to achieve sampling diversity across training episodes, ensuring statistical robustness and generalization of the results.

Regarding the training configuration of HONEST-CAV, the number of parallel worker processes for multi-processing is set to $N=8$, and the maximum number of training episodes/epochs $K$ is 150. For the SAC-based traffic signal control agent, the learning rate, discount factor, batch size, and replay buffer capacity are set to 0.0003, 0.99, 32, and 100{,}000, respectively. 

\subsection{Performance on Benchmark Scenario}
In this section, we compare the conventional Webster SPaT-based TSC method \cite{webster1958traffic}, which serves as the baseline, with the independent reinforcement learning (IRL) approach and the proposed MARL-based global-reward-sharing TSC method. First, the Webster method is a signal control strategy that minimizes the total traffic delay at each individual intersection under static traffic demand. The traffic demand at each intersection is estimated by counting historically observed passing vehicles over a 400-second interval, and the Krauss model is adopted as the car-following model. Second, the IRL algorithm updates each intersection independently using its local reward and a local Q-network. Finally, we integrate the MLTPA with the MARL-based TSC to further enhance overall system performance with the assistance of CAVs, where the CAV penetration rate is set to 60\%.

Network-level performance is evaluated using three metrics: average energy consumption (total energy usage divided by total vehicle travel distance), average idling time (total idling time per vehicle), and average speed (total travel distance divided by total travel time). Each RL-based method is evaluated over 50 Monte Carlo simulations with different random seeds, and the averaged results are summarized in Table~\ref{tab:benchmark}.

\begin{figure}[htbp]
  \centering
    \begin{subfigure}[b]{0.31\linewidth}
    \centering
    \includegraphics[width=1.0\linewidth]{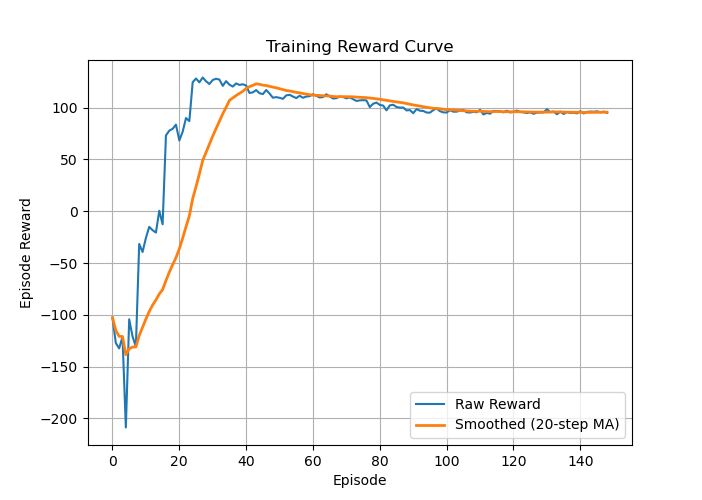}
    \caption{IRL+Krauss}
    \label{fig:IRL_reward}
  \end{subfigure}
  \begin{subfigure}[b]{0.32\linewidth}
    \centering
    \includegraphics[width=1.0\linewidth]{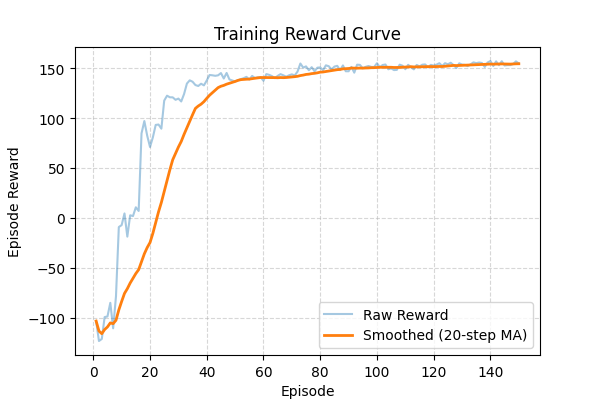}
    \caption{MARL+Krauss}
    \label{fig:MARL_reward}
  \end{subfigure}
  \hfill
  \begin{subfigure}[b]{0.33\linewidth}
    \centering
    \includegraphics[width=1.00\linewidth]{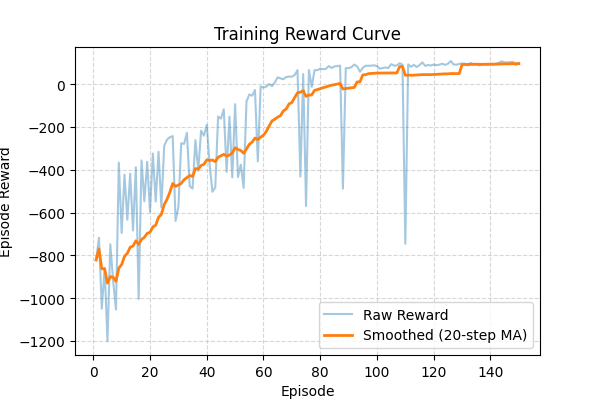}
    \caption{MARL+MLTPA}
    \label{fig:MARL+MLTPA}
  \end{subfigure}
  \caption{The Reward Training Curves}
  \label{fig:fixed_reward}
\end{figure}

The reward training curves for IRL TSC + Krauss, MARL TSC + Krauss and MARL TSC + MLTPA (under a fixed CAV proportion) are presented in Fig.~\ref{fig:fixed_reward}. The reward values converge as training progresses, demonstrating the stability and effectiveness of the proposed HONEST-CAV framework. However, the training reward becomes smaller when MLTPA is added, this is mainly because MLTPA makes traffic smoother but increases the measured queue occupancy and reduces per-cycle throughput, causing lower training reward even though the overall traffic efficiency becomes better.

Overall, the results indicate that jointly optimizing TSC and vehicle trajectory planning leads to significant performance gains over the baseline, including reductions in energy consumption and idling time, as well as improvements in traffic flow efficiency.

\begin{table*}[t]
\caption{Performance Comparison in the Benchmark Scenario}
\centering
\footnotesize
{\scalebox{1.0}{
\begin{tabular}{lcccc}
\toprule
Metric & Baseline & IRL TSC + Krauss & MARL TSC + Krauss & MARL TSC + MLTPA \\
\midrule
Avg. Energy (L/100 km) & 11.71 & 11.32 (-3.33\%)& 10.99 (-6.13\%) & \textbf{10.51 (-10.23\%)} \\
Avg. Idling Time (s/veh) & 49.79 & 47.05 (-5.51\%) & 36.32 (-27.05\%) & \textbf{26.97 (-45.83\%)} \\
Avg. Speed (m/s) & 8.21 & 8.51(+3.52\%) & 8.72 (+6.25\%) & \textbf{8.84 (+7.67\%)}\\
\bottomrule
\end{tabular}
}}
\label{tab:benchmark}
\end{table*}

\subsection{Discussion of CAV Proportions and Powertrain Types}
In this paper, the CAV/CAEV proportion is defined as the ratio of CAVs/CAEVs among all vehicles generated within a fixed network area during the simulation period of each episode. To comprehensively evaluate the performance of HONEST-CAV under varying proportions, we train the model with randomly generated proportion for each episode, and conduct 50 testing experiments using different random seeds, with proportions reasonably distributed across the range of [10\%, 90\%]. Specifically, for energy consumption evaluation, EV energy consumption (kWh) is converted to its gasoline-equivalent (L) based on the lower heating value energy equivalence, using 1 kWh $\approx$ 3.6 MJ $\approx$ 0.112 L of gasoline \cite{center2014fuel}. The MMPEVEM electricity consumption model \cite{koch2021accurate}, a SUMO-compatible EV powertrain model that computes energy consumption and regenerative braking by explicitly modeling powertrain components and operating states.

\begin{figure}[htbp]
  \centering
  \begin{subfigure}[b]{0.48\linewidth}
    \centering
    \includegraphics[width=1.0\linewidth]{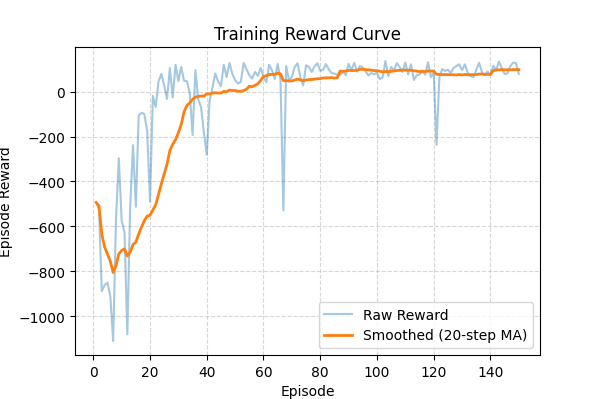}
    \caption{ICEV Environment}
    \label{fig:ICEV}
  \end{subfigure}
  \hfill
  \begin{subfigure}[b]{0.48\linewidth}
    \centering
    \includegraphics[width=1.00\linewidth]{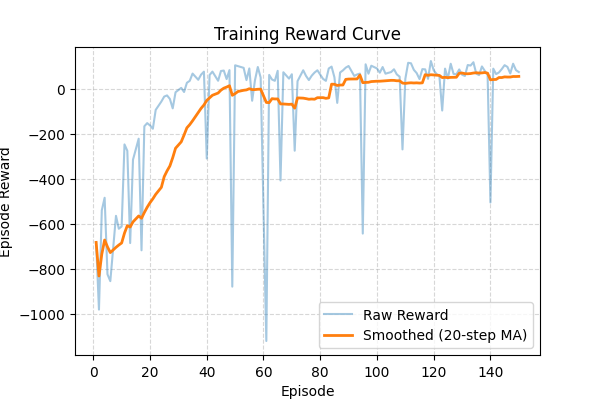}
    \caption{EV Environment}
    \label{fig:EV}
  \end{subfigure}
  \caption{The Reward Training Curves with Random Proportions}
  \label{fig:rand_reward}
\end{figure}

The effective reward training curve for HONEST-CAV and HONEST-CAEV are depicted in Fig.~\ref{fig:rand_reward}.

\begin{figure}[htbp]
  \centering
  \includegraphics[width=\linewidth]{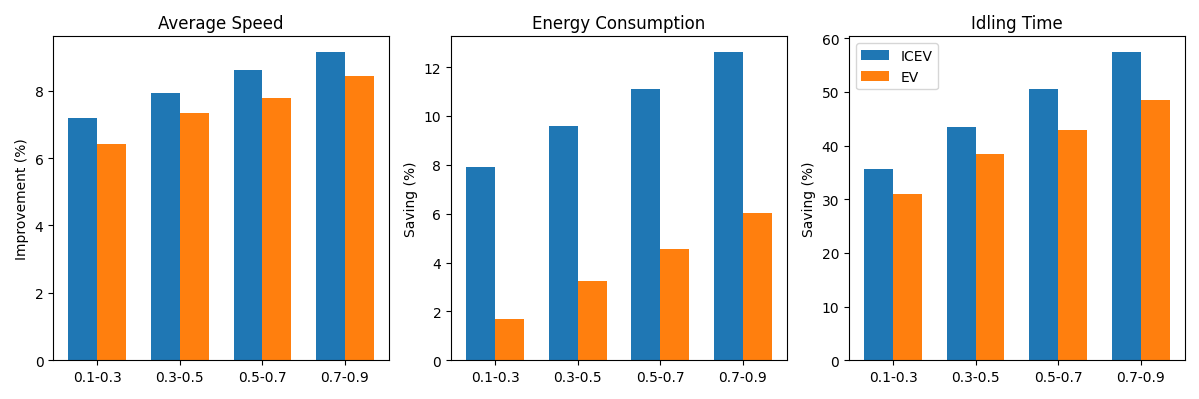}
  \caption{The Savings under Different CAV/CAEV Proportions}
  \label{fig:HONEST}
\end{figure}

Moreover, to quantify HONEST-CAV’s performance across different CAV proportions, we cluster the results from all random seeds into four proportion ranges—[10\%, 30\%), [30\%, 50\%), [50\%, 70\%), and [70\%, 90\%], and average the corresponding performance metrics, as shown in Table~\ref{tab:ICEV} and Fig.~\ref{fig:HONEST}. We observed that performance metrics consistently improve with higher CAV proportions. Compared to the baseline model (Demand-based Fixed TSC + Krauss Model), HONEST-CAV achieves improvements of 9.15\% in average speed, 12.62\% in energy consumption, and 57.47\% in idling time within the [70\%, 90\%] CAV proportion range. These results demonstrate the scalability and robustness of HONEST-CAV across varying CAV proportions and highlight its ability to fully exploit the potential benefits of connected and automated vehicles.

\begin{table*}[t]
\caption{Performance Comparison under Different CAV Proportions}
\centering
\footnotesize
\begin{tabular}{lcccc}
\toprule
Metric & 0.1-0.3 & 0.3-0.5 & 0.5-0.7 & 0.7-0.9 \\
\midrule
Avg. Energy (L/100 km) & 10.79 (-7.88\%) & 10.59 (-9.57\%) & 10.41 (-11.11\%) & \textbf{10.23 (-12.62\%)} \\
Avg. Idling Time (s/veh) & 32.06 (-35.60\%) & 28.17 (-43.42\%) & 24.61 (-50.56\%) & \textbf{21.18 (-57.47\%)} \\
Avg. Speed (m/s) & 8.80 (+7.16\%) & 8.86 (+7.93\%) & 8.92 (+8.62\%) & \textbf{8.96 (+9.15\%)}\\
\bottomrule
\end{tabular}
\label{tab:ICEV}
\end{table*}

In Table~\ref{tab:EV}, the baseline results are obtained under a fully electrified environment. For HONEST-CAEV, similar performance trends are observed as the CAV/CAEV proportion increases; however, the overall energy consumption is consistently lower across all proportion ranges, demonstrating the additional benefits brought by electrification. In particular, compared with the ICEV environment, HONEST-CAEV achieves up to a 90.13\% reduction in energy consumption within the [70\%, 90\%] CAV proportion range.

\begin{table*}[t]
\caption{Performance Comparison under Different CAEV Proportions}
\centering
\footnotesize
\begin{tabular}{lccccc}
\toprule
Metric & Baseline (EV-based) & 0.1-0.3 & 0.3-0.5 & 0.5-0.7 & 0.7-0.9 \\
\midrule
Avg. Energy (L/100 km) & 1.08 & 1.06 (-1.88\%) & 1.04 (-3.43\%) & 1.03 (-4.74\%) & \textbf{1.01 (-6.20\%)} \\
Avg. Idling Time (s/veh) & 49.80 & 34.37 (-30.98\%) & 30.69 (-38.38\%) & 28.48 (-42.82\%) & \textbf{25.63 (-48.53\%)} \\
Avg. Speed (m/s) & 8.21 & 8.74 (+6.41\%) & 8.81 (+7.35\%) & 8.85 (+7.78\%) & \textbf{8.90 (+8.45\%)}\\
\bottomrule
\end{tabular}
\label{tab:EV}
\end{table*}

\section{Conclusion and Future Work}
This paper presents a hierarchical co-optimization framework for vehicle trajectory planning and traffic signal control in large-scale urban networks with CAVs. By integrating a model-free MARL approach for traffic signal control with a learning-based EAD strategy at the vehicle level, the proposed framework effectively bridges real-time signal control with individual vehicle trajectory optimization. A key component of the framework is a green phase prediction algorithm, which enables coordination between signal and vehicle control layers, ensuring the timely generation of smooth and energy-efficient speed profiles for CAVs.

Through comprehensive experiments across various network sizes and CAV proportions, the results demonstrate the robustness, scalability, and effectiveness of the proposed method in improving traffic efficiency and reducing energy consumption. In the benchmark scenario, the proposed HONEST-CAV significantly outperformed the baseline in terms of energy consumption, idling time, average speed and simulation time.

Moreover, HONEST-CAV continues to improve performance as the CAV proportion increases, demonstrating strong scalability and robustness across varying levels of CAV deployment, and effectively harnessing the potential benefits of CAV to enhance overall traffic system efficiency. In addition, with the ICEV replaced to EV, the energy usage is further reduced, which reveals the potential of traffic electrification.

In future work, a zone division approach would be considered for city-level networks, applying the HONEST-CAV framework within each zone. This additional zone-level layer will enable the extension of our method to full-scale urban deployments.

\bibliographystyle{IEEEtran}

\bibliography{main.bib}

\end{document}